\documentclass{article}

\usepackage{times}
\usepackage{graphicx}
\usepackage{subfigure} 
\usepackage{amssymb}
\usepackage{amsthm}
\usepackage{amsmath}
\usepackage{natbib}

\usepackage{algorithm}
\usepackage{algorithmic}

\newtheorem{definition}{Definition}
\usepackage{hyperref}

\usepackage[accepted]{icml2014} 

\icmltitlerunning{Extended Formulations for Online Linear Bandit Optimization}

\begin{document} 

\twocolumn[
\icmltitle{Extended Formulations for Online Linear Bandit Optimization\\ 
           }

\icmlauthor{Shaona Ghosh}{shaona.ghosh@ecs.soton.ac.uk}
\icmladdress{University of Southampton,
            University Road, Southampton, UK}
\icmlauthor{Adam Pr\"{u}gel-Bennett}{apb@ecs.soton.ac.uk}
\icmladdress{University of Southampton,
            University Road, Southampton, UK}

\icmlkeywords{ Online Learning, Online Linear Optimization, Adversarial Multi-arm Bandits}

\vskip 0.3in
]

\begin{abstract} 
On-line linear optimization on combinatorial action sets (d-dimensional actions) with bandit feedback, is known to have complexity in the order of the dimension of the problem. The exponential weighted strategy achieves the best known regret bound that is of the order of $d^{2}\sqrt{n}$ (where $d$ is the dimension of the problem, $n$ is the time horizon ). However, such strategies are provably suboptimal or computationally inefficient. The complexity is attributed to the combinatorial structure of the action set and the dearth of efficient exploration strategies of the set. Mirror descent with entropic regularization function comes close to solving this problem by enforcing a meticulous projection of weights with an inherent boundary condition. Entropic regularization in mirror descent is the only known way of achieving a logarithmic dependence on the dimension. Here, we argue otherwise and recover the original intuition of exponential weighting by borrowing  a technique from discrete optimization and approximation algorithms called `extended formulation'. Such formulations appeal to the underlying geometry of the set with a guaranteed logarithmic dependence on the dimension underpinned by an information theoretic entropic analysis. We show that with such formulations, exponential weighting can achieve logarithmic dependence on the dimension of the set.
\end{abstract} 

\section{Introduction}
\label{intro}
Online linear optimization is a natural generalization of the  the basic adversarial (non-stochastic) or worst case multi-arm bandit framework~\citep{auer2002nonstochastic}, to the domain of convex optimization, where the set of actions is replaced by a compact action set $\mathcal{A}\subset\mathbb{R}^d$ and the loss is a linear function on the action set $\mathcal{A}$.

Despite being a compelling and widely used framework, the problems become challenging to address when the forecaster's decision set is the set of all possible overlapping actions, thus having a combinatorial structure. In this case, the forecaster's action at every round is an element of the combinatorial space. Examples of such problems are maximum weight cut in a graph~\citep{goemans1995improved}, planted clique problem~\citep{jis1979computers} and others. For instance, in the spanning trees with $K$-clique problem, the number of spanning trees of a clique of $K$ elements is exponential in $K$, hence the size of the all possible actions in the action set is also exponential in $K$~\citep{cesa2012combinatorial}. Such combinatorial optimization problems have general tractability issues~\citep{goemans1995improved}. Online linear optimization techniques prove good in such problems when the number of actions $N$ is exponential in the natural parameters of the problem. However, with partial or bandit feedback, such techniques have suboptimal regret bounds. The best known bound is of the order $d^{2}\sqrt{n}$~\citep{bubeck2012towards} with a computationally hard John's exploration technique~\citep{ball1997elementary}. Specifically, in the line of research of online combinatorial optimization, we address the following open questions posed by Bubeck et al. ~\yrcite{bubeck2012regret}. For the combinatorial bandit optimization where actions are d-dimensional, can we have an optimal regret bound with a computationally efficient strategy? Is there a natural way one can characterize the combinatorial action sets for which such optimal regret bounds are obtained under bandit feedback?

We address this question by using a notion from convex optimization that attributes the complexity of the problem in computing a complex set, to an efficient representation of the set. We employ techniques from non-negative and semi-definite linear programming to exploit the natural combinatorial structure of the problem. Particularly, we use a technique called `extended formulation'- a lift and projection technique where the complicated combinatorial convex set (action set) is approximated by a simpler convex set whose linear image is the original convex set~\citep{gouveia2011lifts}. Particularly, we are interested in the approximation of the convex set based on the orthant or semi-definite/non-negative cone that has an associated natural barrier penalty~\citep{nesterov1994interior,chandrasekaran2013computational}.  This technique shows that weaker approximations can lead to improved performance on the combinatorial optimization. Further, such hierarchy of approximations can naturally characterize the action sets for which optimal regret bounds can be attained.

\subsection{Contribution}
Specifically, our contributions are the following:
\begin{itemize}
\item We present the ``extended" exponential weighted algorithm. Our algorithm has an exponential weighting technique for the different actions; the weights in every round form the slack matrix (extended formulation). The minimum non-negative rank-$r$ factorization, $r \ll d$, ($d$ is the dimensionality of the problem) of the slack matrix in every round guides the exploration of the set.  
\item  We introduce an inherent regularity measure by which the actions are weighted, that is a strikingly good indicator of the regret gap. Our method defines slack variables or inequalities given by the linear scalar loss at every round. The notion of non-negative `slack' based regularization, measures the distance of each vertex (in geometric intuition, vertex corresponds to action) in the set from the origin, given the distance from the hyperplane (loss observed) and the vertex played in every iteration. The regularization penalizes for being too far away from either the loss or the last action played for exploitation while revealing boundary information (distance from origin) for exploration.
\item As an interesting consequence of the existence of the slack matrix, we show a minimum rank-$r$; $r \ll d$ sampling technique that guides the exploration and the unbiased loss estimator. In the setting of extended formulations, we analyse the notion of dimension complexity as the measure of complexity of matrices involved.
\item Our theoretical analysis is an indicator that entropic bounds exist beyond mirror descent based regularization. Contrary to the projection of the weights of the actions as in mirror descent, we use a counter-intuitive notion to lift and extend the complex action set itself to higher dimensions, take a simplified linear projection of this lift and work there. We confirm the computational advantage of the extended formulation over the existing exponential weighted techniques through empirical results on simulated and real dataset. 
\end{itemize}

\section{Relation with Previous Work}

Dani et al.~\yrcite{dani2008price} showed that optimal regret bounds of the order of $\mathcal{O}(\sqrt{n})$ (where $n$ is the number of rounds) was first obtained by using a variant of the original adversarial bandit Exp3~\cite{auer1995gambling} strategy. Their strategy explored the set of actions uniformly over a barycentric spanner by selecting actions that are separated by a maximum distance. They also showed that without further assumptions, for $\sqrt{dn\log{\left|\mathcal{A}\right|}}$, the bound is not improvable for $\mathcal{A}={\left[-1,1\right]}^d$. This exploration strategy was later refined by Bianchi et al.~\yrcite{cesa2012combinatorial}, for combinatorial symmetric action sets $\mathcal{\left|{A}\right|}$ by using a bounded $L_\infty$ loss assumption and using a uniform distribution over the actions, while proving that the regret bound of the order $\sqrt{nd\left|\mathcal{A}\right|}$ is not improvable in general for concrete choices of $\mathcal{A}$; but the bound is suboptimal in path planning problem. In both Dani et al.~\yrcite{dani2008price} and Bianchi et al.~\yrcite{cesa2012combinatorial},  the forecaster strategy is based on the probability distribution as in Exp3 by Auer~\yrcite{auer1995gambling}. However, for a set of $k$ arms, Exp3 scales with $\sqrt{nN\ln{n}}$, and for discretised $k$ into $d$ as in the combinatorial setting, $N$ possible actions is exponential in $d$.  This work was completed by Bubeck et al.~\yrcite{bubeck2012towards}, where an optimal exploration using John's theorem~\cite{ball1997elementary} from convex geometry is used on an adaptation of Exp3 called Exp2, to obtain optimal regret bound of $\sqrt{dn\log{\mathcal{A}}}$ for any set of finite actions.

Simultaneously, there are other results in the same direction of research using gradient descent, mirror descent and perturbation approaches~\cite{bubeck2012regret}. Using mirror descent for online linear optimization with bandit feedback on a computationally efficient strategy was provided by ~\cite{abernethy2008competing} using self concordant barriers. This approach enforced a natural barrier based on the local geometry of the convex set to determine the optimal distance of exploration inside the convex hull of the action set. However, this strategy results in a suboptimal dependency on the dimension $d$ of the action set given by $\mathcal{O}(d^2\sqrt{n})$ with bounded scalar loss assumption. This result is improved by Bubeck~\yrcite{bubeck2012towards} using Exp2 strategy to attain $\mathcal{O}(d\sqrt{n})$. However, Audibert~\yrcite{audibert2011minimax} proved that Exp2 is a provably suboptimal strategy in the combinatorial setting. Their work showed that when the action set is combinatorial $\mathcal{A}\subset{\left[0,1\right]}^d$ and loss $\mathcal{L}={\left[0,1\right]}^d$, the minimax regret in the full information and semi-bandit case is of the order $d\sqrt{n}$ while with bandit feedback the order is the sub-optimal $d^{3/2}\sqrt{n}$. The optimal regret bound is also obtained by the mirror descent strategies on the simplex and the Euclidean ball action sets. OSMD for the Euclidean ball in Bubeck~\yrcite{bubeck2012towards},~\cite{ball1997elementary}, achieves regret of the order $\sqrt{dn}$. For more information on combinatorial linear optimization, interested reader may refer~\cite{bubeck2012regret}. Cesa Bianchi et al. have established in~\yrcite{cesa2012mirror}, the unified analysis of mirror descent and fixed share of weights between experts in the online setting. Although, they have showed a logarithmic dependence on the dimension for the generalized case, our work is different in that we capture the loss estimation and show that in cases, the dependency is logarithmic on $r$ where $r \ll d$. However, our analysis of the oblivious adversary is close to their analysis for the generalized adversary on the simplex.

Although, efficient in terms of playing the optimal strategy, the exponentially weighted techniques are computationally and provably sub-optimal given the complexity of the action set and the tools used for performing the exploration. One can question if the non-triviality is in reality much more fundamental and obvious than that. From the knowledge of convex optimization, one knows that the complexity of a convex set is associated to how well the set is represented. A sufficiently complex set can be efficiently represented as a combination of low complexity simpler sets. The added complexity associated with the simplification is offset by the low order complexity of the overall simplified problem. Our work here follows this intuition and line of research.

\subsection{Extended Formulations}
\label{sec:ExtForm}
\begin{figure}[t]
\begin{center}
\subfigure[]{\includegraphics[width=0.45\columnwidth,trim=10mm 25mm 25mm 45mm, clip]{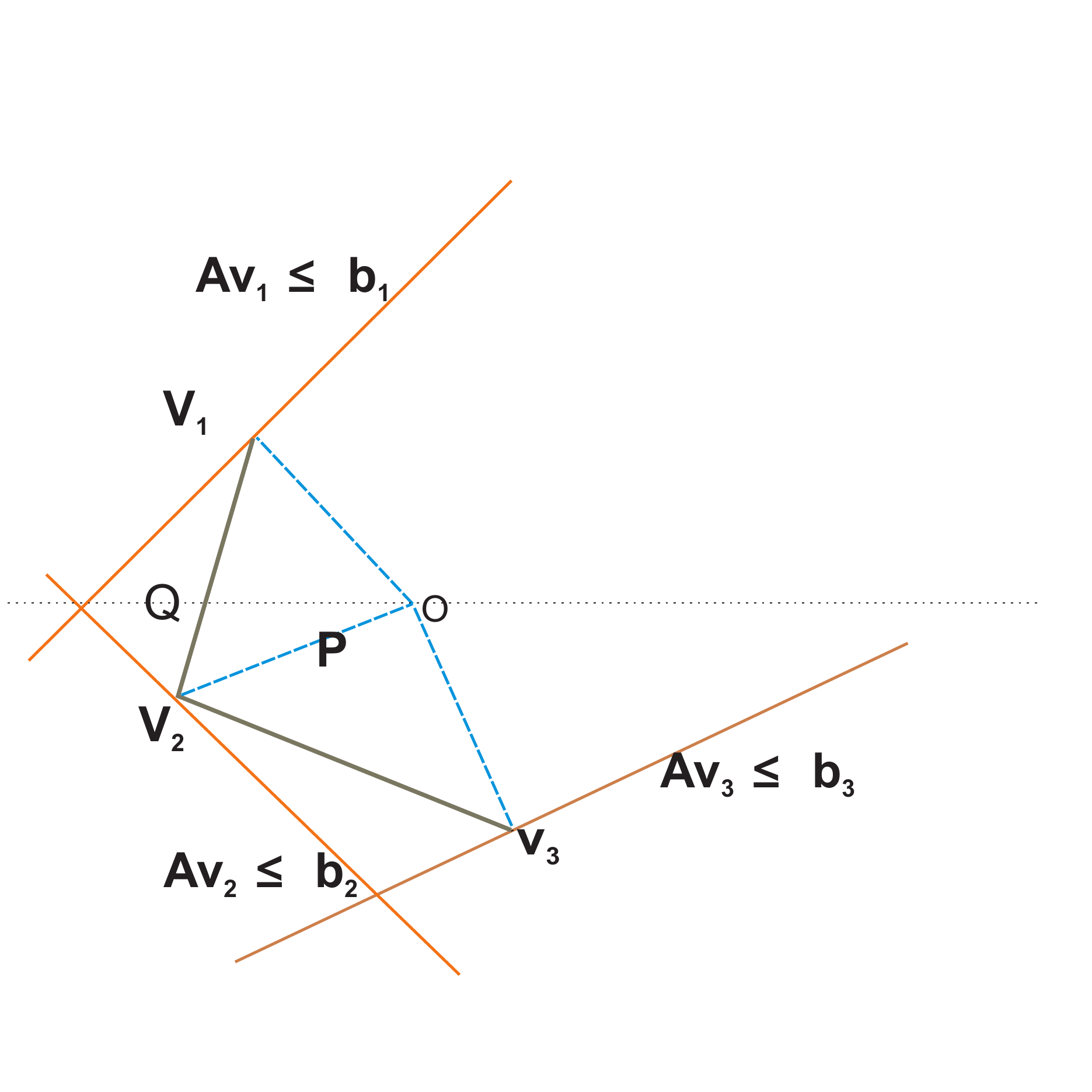}}
\subfigure[]{\includegraphics[width=0.45\columnwidth,trim=10mm 25mm 25mm 25mm, clip]{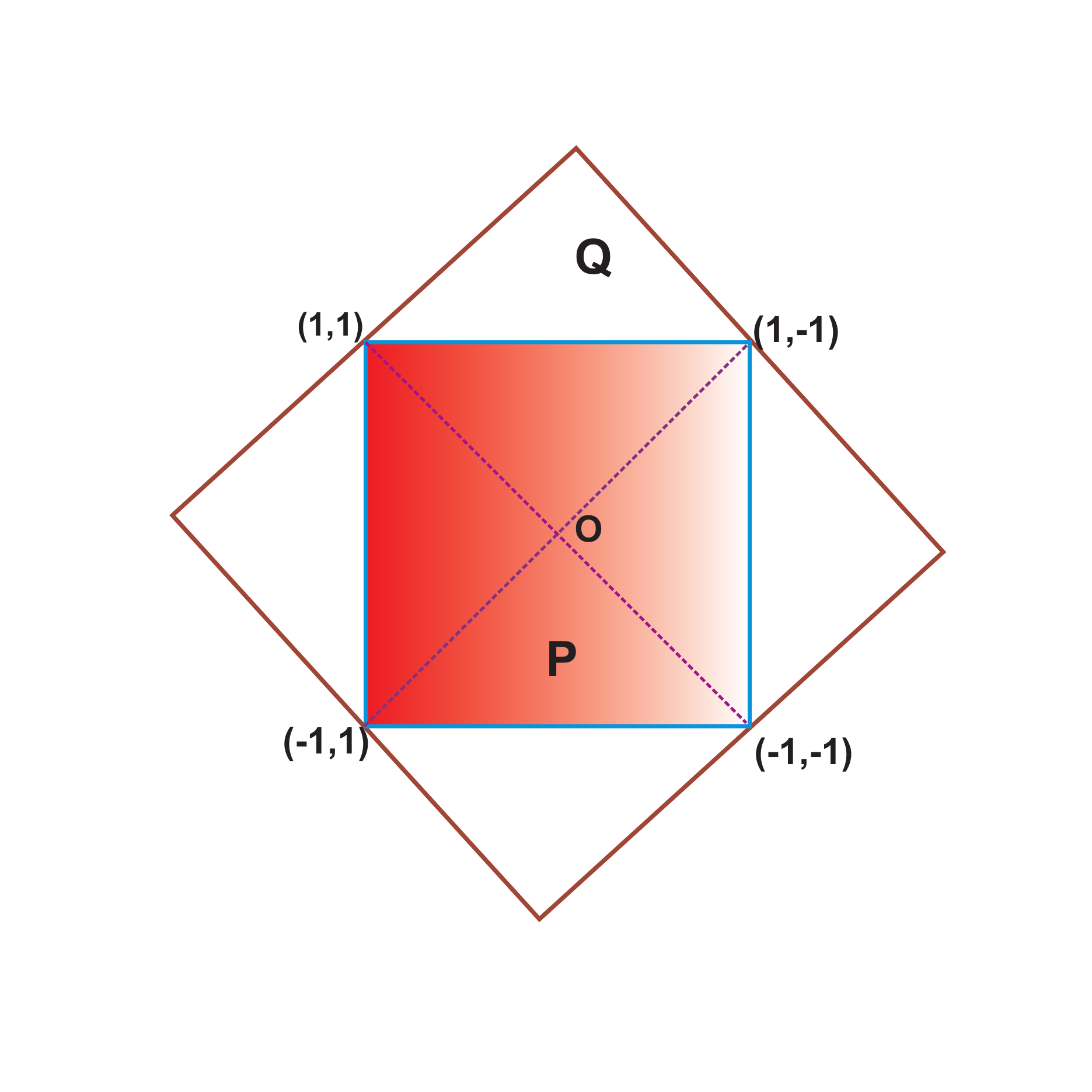}}
\caption{Illustration of extended formulation of convex compact set $P$ (a) $Q$ is the lifted or extended set, both $P$ and $Q$ are in the non-negative orthant cone. Vertices are denoted by $v_i$, $b_i$'s are the distances of the hyperplanes from the origin (b) Toy example of the infinite polytope $P$ and its extended formulation $Q$. Slack measures the distance from the origin $O$ to the vertices.}
\label{fig:ext form}
\end{center}
\end{figure}

An extended formulation~\cite{conforti2010extended} is a way of representing the feasible set of solutions that is naturally exponential in the size of the data, to a formulation polynomial in the natural parameters of the problem by the introduction of a polynomial number of new variables. Where possible, the extended formulation or its approximation, simplifies the computing complexity associated with the problem in the linear programming paradigm. Yannakakis's~\yrcite{yannakakis1991expressing} seminal work naturally characterizes the size of the extended formulation by giving the smallest size of the extension that represents the original set well. The recent work of Gouveia et al.~\yrcite{gouveia2011lifts} and Fiorini et al.~\yrcite{fiorini2012linear} generalizes the theorem from linear programming paradigm for convex optimization. The techniques are also known as lift-and-project~\cite{balas1993lift} and approximation techniques~\cite{chandrasekaran2013computational}. Our work employs this striking technique to the online linear optimization setting to exploit the underlying geometry in the hope of recovering some structure in the combinatorial action set.
In the illustration shown in Figure~\ref{fig:ext form}, $P$ is the original compact set and $Q$ is the extension, such that $P \subseteq Q$.
 
\section{Extended Formulations in Linear Bandits}
\subsection{Problem Setup}
In the formal setting of the combinatorial bandit optimization problem, we consider an adversarial environment with limited feedback. The prediction game as described in the previous sections proceeds in a series of rounds  $t=1,\dots,T$. The action set forms a subset of a hypercube $\mathcal{A}\subset{\left[0,1\right]}^d$ that comprises set of all possible actions, where $d$ is the dimension of the hypercube. When the feasible solutions of a combinatorial optimization problem are encoded as 0/1-points in $\mathbb{R}^d$,  they yield a convex hull polytope of the resulting points. The loss is assumed to be non-negative $\mathcal{L}=\left[0,1\right]^d$. At every  round $t$, the forecaster chooses an action $a_t\in\mathcal{A}$; the action is represented by its incidence vector as corners of the hypercube. The cardinality of all possible actions is denoted by $N$. The adversary secretly selects the loss that is incurred by the forecaster and is defined as $a_t^Tl_t$. The forecaster's strategy at every round is to choose a probability distribution $p_{t-1}\left(1\right),\dots,p_{t-1}\left(N\right)$ over the set of actions such that $p_{t-1}(k)\geq 0$ for all $k = 1,\dots,N$. Note that $\sum_{k=1}^Np_{t-1}(k)=1$, and action $a_t=k$ is drawn with probability $p_{t-1}(k)$. The objective of the forecaster is to minimize the pseudo regret defined in Equation~\ref{eq:regret}. The expectation in~\ref{eq:regret} is with respect to the forecaster's internal randomization and possible adversarial randomization. Note that this problem formulation is identical to the adversarial bandit framework when $d=N$ and the action selected at each round $a_t$ forms the canonical basis.
\begin{equation}
\label{eq:regret}
\overline{R_T} = \mathbb{E}\sum_{t=1}^{T}a_t^{T}z_t - \min_{a\in\mathcal{A}}\mathbb{E}\sum_{t=1}^{T}a^{T}z_t
\end{equation}

\subsection{Slack based Regularization}
\begin{definition}
Let $P = conv\left\lbrace{{a}|a\in \mathcal{A} \subset {\left\lbrace{\mathbb{R}}\right\rbrace}^d}\right\rbrace$ be the convex hull of the action set with a non-empty interior, then for every loss estimate sampled randomly from the simplex $l \in {\left\lbrace{0,1}\right\rbrace}^d$ by the adversary, there is an extended set $Q = \lbrace x | Ax \leq l \rbrace$, where $x$ is the action played, $A$ is the action set, $l$ is the loss observed. Both $P$ and $Q$ are defined on the positive orthant.
\end{definition}
The above definition implies that, if $P$ is the convex hull of all actions defined on the positive quadrant, let there be a set of hyperplanes given by the loss observed at every round; then $Q$ defines a set of inequalities that define on which side of the hyperplanes defined by the losses, the vertices lie. 
\begin{definition}
\label{def:slackreg}
The slack regularity for all action $a \in \mathcal{A}$, is the non-negative measure of how far the action is from the origin of the set $P$. It is the difference of the distance of the hyperplane from the origin defined by the loss observed $l$ proportional to the distance of the action from the last action played:$\left|{a_t^{T}{l}_t-a_t^{T}a}\right|$, where $a_t^{T}{l}_t$ defines the hyperplane for round $t$, $a_t$ is the action played.
\end{definition}
The slack regularity measure is the linear projection between $P$ and $Q$, and defines a much simpler slack set called the slack matrix. The idea is the original complex set $P$ is extended to $Q$ by introducing inequalities and slack variables - the hyperplanes, then a linear projection technique is used to project the lift back. The linear projection defines the slack matrix. 
\begin{definition}
The non-negative slack matrix defined by the slack regularity measure is the measure of how much any particular action is breaking the inequality or is far from the hyperplane proportional to the action played and is given by $M_{t,i} = \left|{a_t^{T}l_t - a_t^{T}a_i}\right| $ for round $t$ is a $\left({t+1}\right) \times N$ matrix
\end{definition}

\section{Algorithms and Results}
\begin{algorithm}[tb]
   \caption{Extended Exp}
   \label{alg:algo1}
\begin{algorithmic}
   \STATE {\bfseries Input:} learning rate $\eta \ge 0$, mixing coefficient $\alpha \ge 0 $, action set dimensional rank $\delta = \rho\left({\mathcal{A}}\right)$. 
   \REPEAT
   \STATE Initialize $p_{1} = \left({\frac{1}{\delta},\dots,\frac{1}{\delta}}\right)\in \mathbb{R}^{\left|{\mathbb{A}}\right|}$. Let non-negative rank $r = \delta$
   \FOR{$t=1$ {\bfseries to} $T$}
   \STATE $p_{i,t} = \frac{\alpha}{r} + \left({1-\alpha}\right)w_{i,t}$
   \STATE Play action $a_t$ from $p_t$
   \STATE Observe loss $a_t^Tl_t$
   \STATE Update loss ${w_{i,t+1}}=  \frac{p_{i,t}e^{-\eta\left|{a_t^{T}l_t-a_t^{T}a_{i,t}}\right|}}{\sum_{j=1}^{N}p_{j,t}e^{-\eta\left|{a_t^{T}l_t-a_t^{T}a_{j,t}}\right|}}$  
   \STATE $M_{t+1} = [w_{t,i}]_{{1 \le t \leq t+1},{1 \le i \leq N}}$
   \STATE Find minimum non-negative rank $r$ such that $M = \sum_{k=1}^{r}T^kU_k$
   \ENDFOR
   \UNTIL{Time horizon T or no regret}
\end{algorithmic}
\end{algorithm}
 
\begin{figure}[t]
\begin{center}
\centerline{\includegraphics[width=\columnwidth, trim = 45mm 180mm 35mm 85mm, clip]{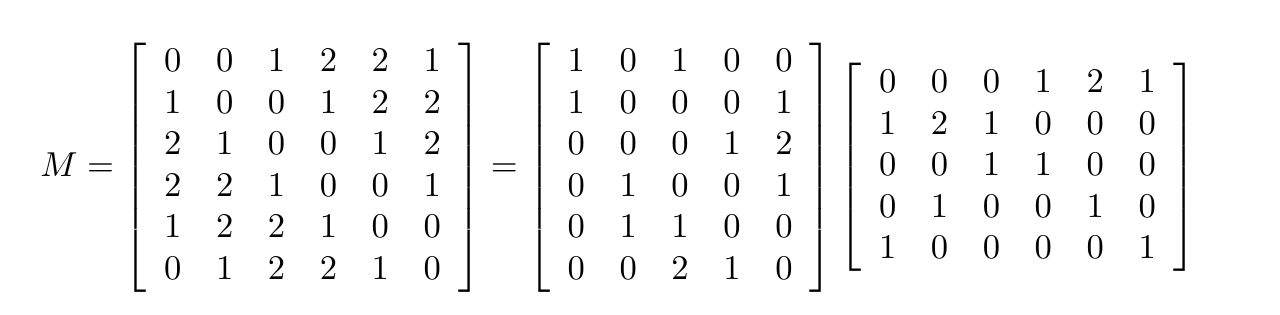}}
\caption{Slack matrix factorization. Slack matrix for the regular hexagon $M$ factored into non-negative matrices. Example taken from~\cite{gouveia2011lifts}. }
\label{icml-historical}
\end{center}
\end{figure}

\begin{algorithm}[tb]
   \caption{Extended Exp2 with Sampling}
   \label{alg:algo2}
\begin{algorithmic}
   \STATE {\bfseries Input:} learning rate $\eta \ge 0$, mixing coefficient $\alpha \ge 0 $, action set dimensional rank $\delta = \rho\left({\mathcal{A}}\right)$. 
   \REPEAT
   \STATE Initialize $p_{1} = \left({\frac{1}{\delta},\dots,\frac{1}{\delta}}\right)\in \mathbb{R}^{\left|{\mathbb{A}}\right|}$. Let non-negative rank $r = \delta$. Let initial covariance matrix $P_1 = cov\left({\mathcal{A}}\right)$
   \FOR{$t=1$ {\bfseries to} $T$}
   \STATE $p_{i,t} = \frac{\alpha}{r}\gamma_{r_{i},t} + \left({1-\alpha}\right)w_{i,t}$
   \STATE Play action $a_t$ from $p_t$
   \STATE Observe loss $a_t^Tl_t$
   \STATE Estimate loss $\tilde{l}_t = P_t^{-1}a_ta_t^{T}l_t$
   \STATE Update loss ${w_{i,t+1}}=  \frac{p_{i,t}e^{-\eta\left|{a_t^{T}\tilde{l}_t-a_t^{T}a_{i,t}}\right|}}{\sum_{j=1}^{N}p_{j,t}e^{-\eta\left|{a_t^{T}\tilde{l}_t-a_t^{T}a_{j,t}}\right|}}$  
   \STATE $M_{t+1} = [w_{t,i}]_{{1 \le t \leq t+1},{1 \le i \leq N}}$
   \STATE Find $r$ non-negative rank-1 matrices $P_k$ such that $M = \sum_{k=1}^{r}P_k$
   \STATE Sample index $r_i$ uniformly $\forall r_i \in \left[{r}\right]$ 
   \STATE Let $\gamma_{r_{i},t} = \frac{P_{r_i},t}{\sum_{r_{i}=1}^{r}P_{r_{i},t}}$
   \ENDFOR
   \UNTIL{Time horizon T or no regret}
\end{algorithmic}
\end{algorithm}
In Algorithm~\ref{alg:algo1}, an exponentially weighted technique is used to weight the actions based on the slack regularity measure defined in Definition ~\ref{def:slackreg}. This defines the weight matrix or the slack matrix $M_{t+1}$ at each round that has a positive rank.
\begin{definition}
\label{def:nnmf}
As long there is a positive rank for the weights matrix $M_t$, there is a guaranteed non-negative factorization possible such that $M = TU$, where $T$ and $U$ are non-negative factors. In general, there is a minimum $k$, such that $M_{i,j} = \sum_{k}T_{i,k}U_{k,j}$
\end{definition}
The Figure~\ref{icml-historical} illustrates the non-negative factorization of the slack matrix.
\subsection{Sampling of low rank approximations }
Definition~\ref{def:nnmf} directly leads to the technique of low rank approximations of the action set. Typically, in the linear optimization algorithms, the crucial step is the loss estimation $\tilde{l}_t$. For the loss estimator to be an unbiased estimator, the actions correlation matrix is generally used. Here, instead of using the whole correlation matrix, we uniformly sample rank-one matrices  from the distribution of $r$ rank-one matrices. 
\begin{definition}
\label{def:sampling}
Given the existence of a positive rank slack weight matrix $M_t$ at every round $t$, that has a minimum $r$ non-negative factorization possible, then $M$ is the sum of $r$ non-negative rank-one matrices as $M = \sum_{r_i=1}^r P_{r_i}$, where $P_{r_i}$ is the rank-1 matrix. At every round, the index $r_i$ is sampled from the distribution $\gamma_{r_i,t}$ uniformly. This index thus sampled generates a subset $S \subset \mathcal{A}$ for actions in $S$ to be explored also such that the rank-$\left|{S}\right|$ matrix $P_{r_i,t} = \sum_{a \in S}w_t\left({a}\right)aa^T$
\end{definition}
It is important to note here that unlike in earlier approaches, we reduce the complexity of exploration by sampling from $\gamma_{r_i,t}$ that gives the bound that depends on the logarithm of $r$ in our analysis, where $r \ll d$. Another important point we would like to make here is that it can be shown that the $P_{r_i,t}$'s  compute the slack matrix $M$ in expectation (see \citep{conforti2011extended}) over the randomization of the player and the adversary.
\subsection{Complexity of Extension}
It turns out that the non-negative rank of the slack matrix $M$ is the extension complexity of the polytope (hypercube in our case) of actions $P$. In reality, the non negative rank of a slack matrix $S$ of the order $\log_2{d}$ is the extension complexity for dimensionality $d$. The rank of the slack matrix provides lower bounds to the complexity of the extension.  The lower bound on the non-negative rank for a regular $n$-gon that has a $\mathbb{R}_{+}^r$ lift  is given by $r=\mathcal{O}\left(\log_2\left(d\right)\right)$~\cite{gouveia2011lifts}. Similarly, the minimum $r$-lift representation of the combinatorial $d$-dimensional hypercube is bounded by $2d$. 
\subsection{Information Theoretic - Entropic treatment}
Although, we are in the exponential weights setting unlike the mirror descent algorithms, where an entropic regularization function is carefully selected to impose the boundary penalty, in our case, the regularization is inherent in the treatment of the problem. It has been shown that a randomized communication protocol can give a stronger lower bound in the order of base-2 logarithm of the non-negative rank of the slack matrix  (communication matrix) computed in expectation~\cite{faenza2012extended}. Moreover, the uniform sampling of the non-negative rank index $r_i$ in our case, has an associated entropy $H(r_i) = \log \left|S\right|$. In fact, the uniform sampling of the low rank matrices $P_{r_i}$ with bias $w_t$ has an entropy $H(w_t) = w_t \log \frac{1}{w_t} + (1-w_t) \log (1-w_t)$ which is similar to the entropic function in the mirror descent case.
\section{Experiments}
We compared the extended exponential weighted approach with the state-of-the-art exponential weighted algorithms in the adversarial linear optimization bandit setting. 
\subsection{Simulations}
In the first experiment, among a $d$-dimensional network of routes, the optimal route should be selected by the learning algorithm. Typically, we choose $d$ to vary between $10$ and $15$. The environment is an oblivious adversary to the player's actions, simulated to choose fixed but unknown losses at each round.  Losses are bounded and in the range $[0,1]$. The learning algorithm is executed using our basic Extended Exp algorithm. Each action is represented a $d$-dimension incidence vector, with $1$ indicating if an edge or path is present in the route or $0$ otherwise.
\begin{figure*}[ht]
\begin{center}
\subfigure[]{\includegraphics[width=0.85\columnwidth,trim = 35mm 95mm 35mm 95mm, clip]{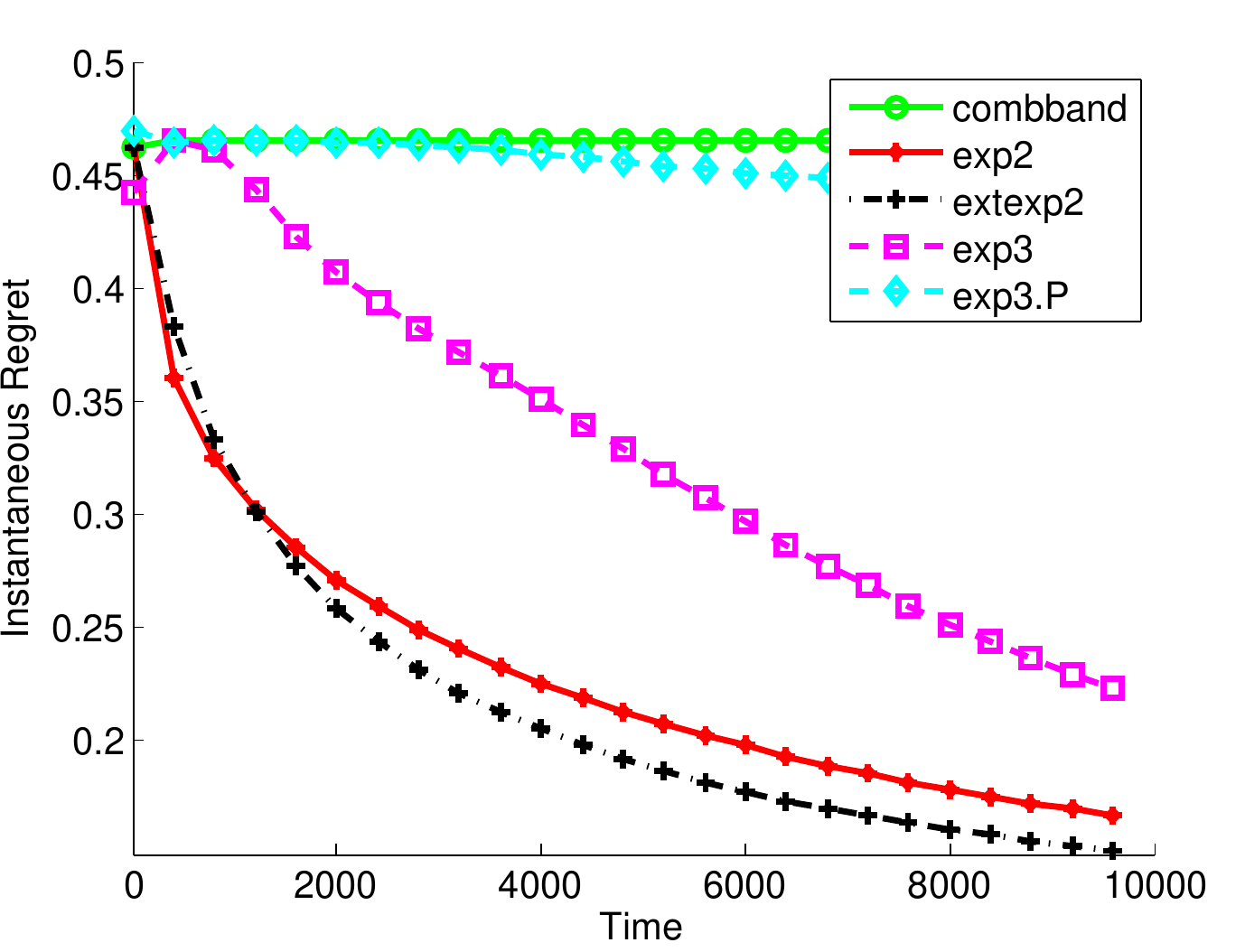}}
\subfigure[]{\includegraphics[width=0.80\columnwidth,trim = 15mm 75mm 23mm 65mm, clip]{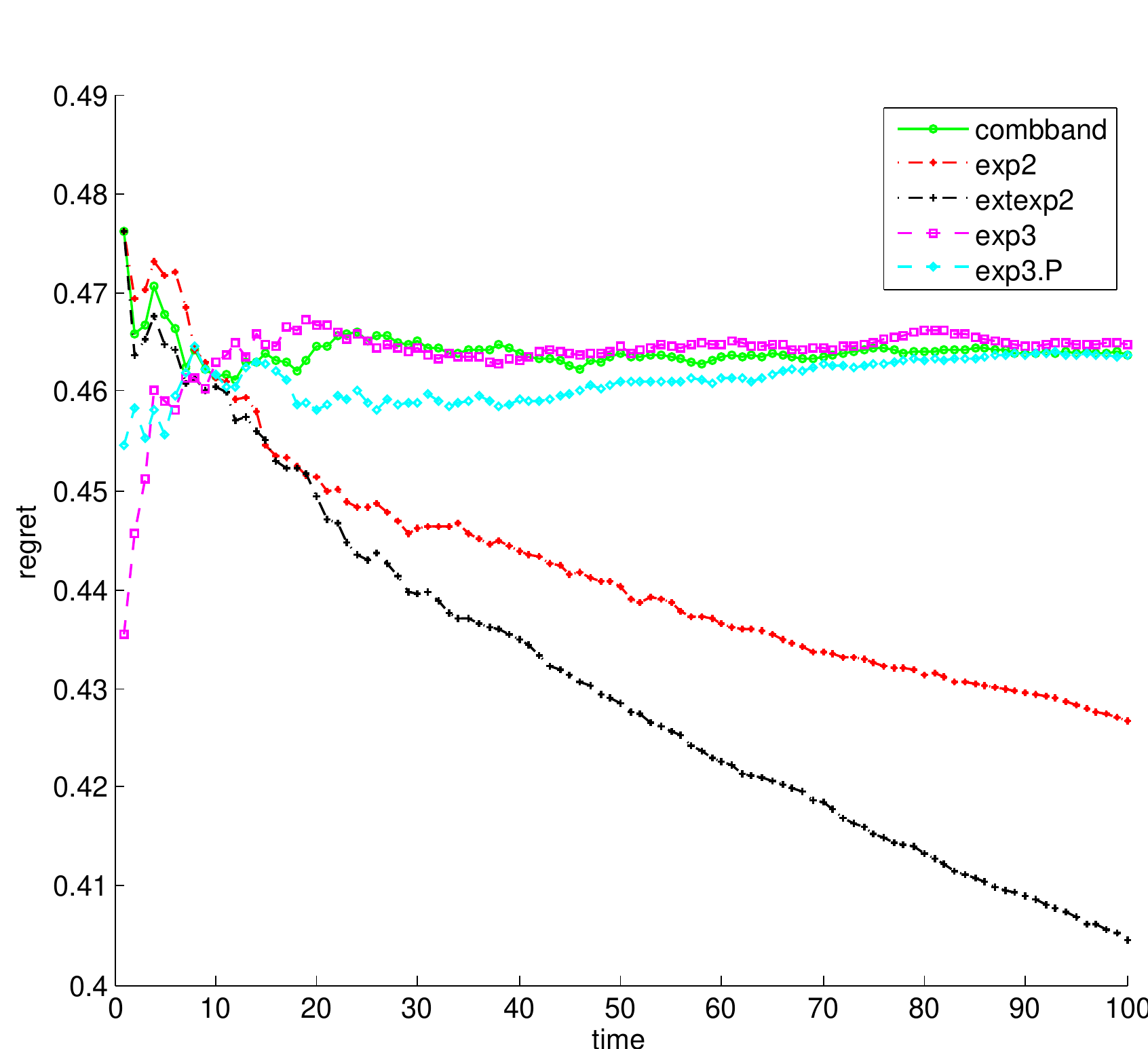}}
\caption{Simulation results with the network dataset with 10 dimensions. Results averaged over 100 runs. Extended Exp2(BLACK) beats the baseline Exp2 (RED) (a) Network dataset with $10$ dimensions (b) Network dataset with $15$ dimensions.}
\label{fig:res_sim10dims}
\end{center}
\end{figure*} 
Figure~\ref{fig:res_sim10dims} displays the results of the cases where $d = 10, 15$ and $d=20$. The performance measure is the instantaneous regret over time; we use psuedo regret here. The number of iterations in the game are $10000$ for $d=10$ and $100$ for $d=15$. In both cases, the results are averaged over $100$ runs of the games. We implemented Algorithm~\ref{alg:algo1}, Exp2~\cite{bubeck2012regret},  Exp3{~\cite{auer2002nonstochastic}, Exp3.P~\cite{auer2002nonstochastic}, and CombBand~\cite{cesa2012combinatorial}. Our baseline is Exp2 which has the best known performance but provably sub-optimal. We could not compare with Exp2 with John's exploration~\cite{bubeck2012towards} as the authors state its computational inefficiency. All the experiments are implemented using Matlab on a notebook with i7-2820QM CPU 2.30 GHz with 8 GB RAM. We see that Extended Exp 2 clearly beats the baseline comfortably against the oblivious adversary. We repeated the experiments with different configurations of the network and different dimensions. Each time, the complexity of the problem increases exponentially with the dimension. Extended Exp2 performs best in all our experiments, the results of the other trials are excluded for brevity.
\subsection{Empirical Datasets}
The dataset is the Jester Online Joke Recommendation dataset~\cite{jester} from the University of Berkeley, which is data collected from 24,983 users with ratings on 36 or more jokes. We consider the ratings of 24,983 users on $20$ jokes, constituting the dense matrix. The ratings vary in the range $[-10.00,10.00]$, including not rated jokes. We scale and normalize the ratings in the range $[0,1]$. For the purpose of our problem, each user represents a $d$ dimensional decision problem, where $d=20$. We do not make the ratings available to the algorithm, instead the ratings are provided by the environment based on the user. In other words, we assume the non-oblivious setting for the adversary, the loss changes with the user or action selected. The goal of the algorithm is to be able to identify the user who has rated the worst on all the jokes on average. As before we execute all the instances of the common exponential weighted algorithms and the basic version of Extended Exp2.  
\begin{figure}[ht]
\begin{center}
\centerline{\includegraphics[width=0.75\columnwidth, trim = 35mm 95mm 35mm 98mm, clip]{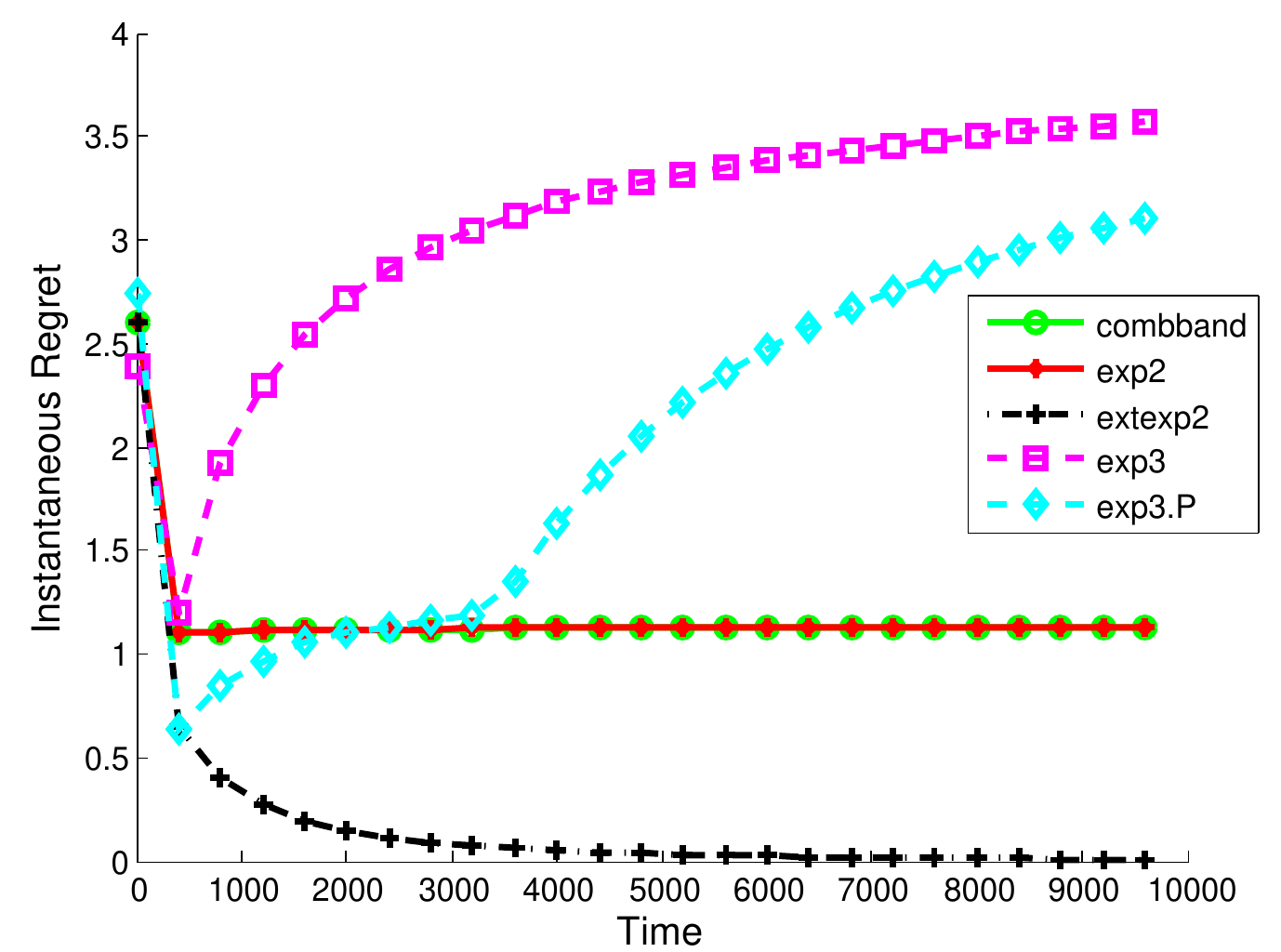}}
\caption{Dataset results with the Jester Online Joke Recommendation dataset with 20 dimensions. Results averaged over 100 runs. Extended Exp2(BLACK) beats the baseline Exp2 (RED).}
\label{fig:jestet20dims}
\end{center}
\end{figure} 
Figure~\ref{fig:jestet20dims} displays the result on the dataset, all the plots  are averaged over $100$ runs of the games. We run each game for $10000$ rounds. All the experiments are implemented on Matlab on a notebook with i7-2820QM CPU 2.30 GHz with 8 GB RAM. We observe that once again Extended Exp2 beats all the others. Quite surprisingly, Exp2 and Combband seem to perform equivalently in the non-oblivious setting. It will be interesting to compare with mirror descent based linear optimization, that is for future work.
\section{Conclusion and Future Work}
We have shown a novel slack based regularization approach to the online linear optimization with bandit information in the exponential weighted setting. We have proved a logarithmic dependence on the dimension complexity of the problem. We showed that in the exponential weighted setting, logarithmic dependence is achievable by a clever regularization, that measures how far an action in the set is, given the loss and the other actions using a trick called 'extended formulation'. As future work, we would like to derive the lower bounds and prove that our results are unimprovable in general. We would also like to investigate how the extended formulation characterizes different action sets. Further, it would be interesting to derive an information theoretic entropic analysis of our method.
\bibliography{ExtendedReferences}
\bibliographystyle{icml2014}
\end{document}